\pdfoutput=1

\documentclass{article}

\usepackage{microtype}
\usepackage{graphicx}
\usepackage{subfigure}

\usepackage{booktabs}
\usepackage{multirow}
\usepackage{makecell}
\usepackage{rotating}

\usepackage{hyperref}

\usepackage[accepted]{icml2023}

\definecolor{periwinkle}{HTML}{7977B8}

\newcommand{\bw}[1]{\textbf{#1}}

\SetKwInOut{Intermediate}{Intermediate}

\usepackage{amsmath}

\usepackage{amssymb}
\usepackage{mathtools}
\usepackage{amsthm}

\usepackage[capitalize,noabbrev]{cleveref}

\theoremstyle{plain}

\theoremstyle{definition}

\theoremstyle{remark}

\usepackage[textsize=tiny]{todonotes}

\icmltitlerunning{
    Exponentially Faster Language Modeling
}

\begin{document}

\twocolumn[
\icmltitle{
    Exponentially Faster Language Modeling
}



\icmlsetsymbol{equal}{*}

\begin{icmlauthorlist}
\icmlauthor{Peter Belcak and Roger Wattenhofer}{yyy}
\end{icmlauthorlist}

\icmlaffiliation{yyy}{Department of XXX, University of YYY, Location, Country}

\icmlcorrespondingauthor{Peter Belcak}{first1.last1@xxx.edu}

\icmlkeywords{Machine Learning, Conditional Execution, Fast Feedforward Networks}

\vskip 0.3in
]




\begin{abstract}
Language models only really need to use an exponential fraction of their neurons for individual inferences.

As proof, we present UltraFastBERT, a BERT variant that uses 0.3\% of its neurons during inference while performing on par with similar BERT models.
UltraFastBERT selectively engages just 12 out of 4095 neurons for each layer inference.
This is achieved by replacing feedforward networks with fast feedforward networks (FFFs).

While no truly efficient implementation currently exists to unlock the full acceleration potential of conditional neural execution, we provide high-level CPU code achieving 78x speedup over the optimized baseline feedforward implementation, and a PyTorch implementation delivering 40x speedup over the equivalent batched feedforward inference.

We publish our training code, benchmarking setup, and model weights.\footnote{\texttt{https://github.com/pbelcak/UltraFastBERT}}
\end{abstract}

\section{Introduction}
\label{section:introduction}
Feedforward layers hold the majority of the parameters of large language models \citep{brown2020language,anil2023palm}.
However, not all of their neurons need to be engaged in the computation of the feedforward layer output at inference time for every input.

For a generally accessible proof, we present UltraFastBERT, a variant of the BERT architecture \citep{devlin2018bert} that replaces feedforward layers with fast feedforward networks.
In terms of downstream performance, UltraFastBERT performs on par with other BERT-like models that are similar in size and undergo similar training procedures.
The intermediate layers of UltraFastBERT are, however, exponentially faster by design: given a feedforward (FF) and a fast feedforward (FFF) network, each with $n$ neurons, the time complexity of a forward pass through the FFF is $\mathcal{O}\left(\log_2 n \right)$ instead of $\mathcal{O}\left(n\right)$ as for FF.
This is a consequence of the fact that FFFs organize their neurons into a balanced binary tree, and execute only one branch of the tree conditionally on the input.

Performing inference on an FFF amounts to performing conditional matrix multiplication (CMM), in which the rows of the input dot with the columns of neural weights one at a time, and the weight column to proceed with is chosen depending on the output of the previous dot-product operation.
In this manner, all neurons are used only by some inputs and no input needs more than just a handful of neurons to be handled by the network.
This is in contrast with dense matrix multiplication (DMM), which lies at the heart of the traditional feedforward networks, and which computes the dot products of all rows with all columns.

No native, efficient implementation of conditional matrix multiplication exists, and no popular deep learning framework offers any interface that could be used to implement it besides a high-level simulation.
We therefore provide a set of CPU implementations based on pointer-batched matrix multiplication routines of the BLAS library.
In a later section, we give a comparison between CPU and GPU implementations at various levels of optimization and note that while there already is clear evidence of significant acceleration, there is potential for more.

\paragraph{The role of attention.}
A large body of literature already addresses the topic of speeding up the execution of the attention mechanism.
We note that for a BERT-base-sized model with the usual pre-training context size of 128 \citep{devlin2018bert}, the per-token inference cost of its attention to all other tokens amounts to only a little more than the cost of 128-neuron feedforward network inference.
We therefore leave the attention layers untouched and focus solely on the intermediate layers hosting the feedforward networks.

\paragraph{Points of comparison.}
BERT-base feedforward networks consist of 3072 neurons.
This is not close to any power of two, and so in the design of UltraFastBERT, we round this number to 4095 -- the number of nodes in a balanced binary tree of maximum depth 11.
In this frame of reference, UltraFastBERT uses only 1/256 (0.04\%) of the 3072 BERT-base neurons for inference. Nevertheless, UltraFastBERT iself consists of 4095 neurons, and so uses 1/341 (0.03\%) of its neurons for inference.

When reporting model performance on downstream tasks in \Cref{section:downstream_performance}, we give both a 3072-neuron and a 4095-neuron baseline for completeness.

\paragraph{Why only 78x and not 341x speedup?}
Dense matrix multiplication is the most optimized mathematical operation in the history of computing.
A tremendous effort has been put into designing memories, chips, instruction sets, and software routines that execute it as fast as possible.
Many of these advancements have been -- be it for their complexity or for competitive advantage --  kept confidential and exposed to the end user only through powerful but restrictive programming interfaces.

Therefore, despite having no need for new hardware, we are still forced to rely on combining high-level linear-algebraic routines to implement CMM, hence the reduction in the speedup.
We elaborate on this in \Cref{section:inference}.

\paragraph{Reproducibility.}
We share the weights of our best model.
While we do not provide an efficient PyTorch or TensorFlow implementation of CMM, the fact that only 12 neurons are used in the inference of UltraFastBERT can be verified simply by masking out the output of all but the chosen neurons, and we give the code for this.

\paragraph{Takeaways.}
\begin{itemize}
    \item We present UltraFastBERT, a BERT-like model that has 4095 neurons but selectively uses only 12 (0.03\%) for inference.
    \item We finetune UltraFastBERT for standard downstream tasks and find that it performs on par with its BERT peers.
    \item We provide a naive implementation of the conditional matrix multiplication that underlies fast feedforward network inference. We find that it leads to a 78x speedup over the natively optimized dense matrix multiplication.

    \item Through UltraFastBERT and the already considerable speedups by simple FFF implementations, we demonstrate the considerable potential of conditional neural execution in language modelling.
\end{itemize}

\begin{table*}[h!]
\centering
\scalebox{0.930}{
    \begin{tabular}{l|cr|ccccccc|c|c|c}
    
    \toprule

    Model &
    \multirow{1}{*}{$N_{\text{T}}$} &
    \multirow{1}{*}{$N_{\text{I}} / N_{\text{T}}$} &
    \multirow{1}{*}{RTE} &
    \multirow{1}{*}{MRPC} &
    \multirow{1}{*}{STSB} &
    \multirow{1}{*}{SST-2} &
    \multirow{1}{*}{MNLI} &
    \multirow{1}{*}{QNLI} &
    \multirow{1}{*}{QQP} &
    \multirow{1}{*}{Avg} &
    \multirow{1}{*}{CoLA} &
    \multirow{1}{*}{Avg} \\

    \midrule
        \multicolumn{12}{l}{\textbf{Baselines}} \\
    \midrule
        \,\,crammedBERT-3072 & 4095 & 100.0\% & 58.8 & 87.6 & 85.2 & 91.9 & 82.8 & 90.4 & 89.0 & 83.6 & 45.0 & 79.3  \\ 
        \,\,crammedBERT-4095 & 3072 & 100.0\% & 57.6 & 89.1 & 85.9 & 91.9 & 81.3 & 90.9 & 87.6 & 83.2 & 47.9 & 79.3 \\ 

    \midrule
        \multicolumn{12}{l}{\textbf{UltraFastBERTs}} \\
    \midrule
        \,\,UltraFastBERT-3072x0 & 3072 & 100.0\% & 56.7 & 88.9 & 86.3 & 92.3 & \bw{82.9} & \bw{92.3} & 88.0 & \bw{83.8} & \bw{48.4} & \bw{79.9}  \\
        \,\,UltraFastBERT-1536x1 & 4608 & 66.6\% & 55.2 & \bw{89.4} & 85.0 & 91.9 & 82.2 & 90.1 & 89.0 & 83.1 & 47.5 & 79.2  \\
        \,\,UltraFastBERT-512x2 & 3584 & 42.9\% & 59.2 & 87.7 & 86.0 & 89.9 & 81.9 & 90.3 & 89.3 & 83.3 & 46.2 & 79.2  \\
        \,\,UltraFastBERT-256x3 & 3840 & 26.7\% & 54.2 & 87.4 & 85.9 & 91.6 & 81.6 & 90.0 & 89.1 & 82.7 & 48.0 & 78.8  \\
        \,\,UltraFastBERT-128x4 & 3968 & 16.1\% & 58.4 & 87.5 & \bw{87.2} & \bw{92.3} & 81.2 & 89.9 & \bw{90.0} & 83.5 & 45.9 & 79.3  \\
        \,\,UltraFastBERT-64x5 & 4032 & 9.5\% & 55.7 & 89.0 & \bw{87.2} & 91.4 & 81.6 & 90.2 & 89.4 & 83.3 & 46.1 & 79.1  \\
        \,\,UltraFastBERT-32x6 & 4064 & 5.5\% & 57.6 & 88.2 & 86.1 & 91.2 & 81.0 & 89.2 & 88.3 & 82.8 & 40.6 & 78.1  \\
        \,\,UltraFastBERT-16x7 & 4080 & 3.1\% & 55.5 & 89.0 & 86.7 & 88.9 & 80.1 & 89.4 & 86.9 & 82.1 & 41.5 & 77.6  \\
        \,\,UltraFastBERT-8x8  & 4088 & 1.8\% & 56.2 & 88.4 & 85.4 & 88.7 & 80.6 & 89.3 & 86.4 & 81.9 & 32.7 & 76.5  \\
        \,\,UltraFastBERT-4x9  & 4092 & 1.0\% & 53.8 & 85.9 & 85.7 & 89.6 & 81.9 & 89.3 & 88.0 & 82.0 & 31.8 & 76.4  \\
        \,\,UltraFastBERT-2x10 & 4094 & 0.5\% & \bw{59.9} & 88.8 & 85.3 & 87.4 & 79.9 & 89.2 & 86.1 & 82.0 & 35.4 & 76.9  \\
        \,\,UltraFastBERT-1x11 & \bw{4095} & \bw{0.3\%} & 57.8 & 88.1 & 86.1 & 89.7 & 80.2 & 89.3 & 87.1 & 82.3 & 37.1 & 77.3 \\
    \midrule
        \multicolumn{12}{l}{\textbf{Final Model}} \\
    \midrule
        \,\,UltraFastBERT-1x11-long  & 4095 & 0.3\% & 60.7 & 87.5 & 86.4 & 89.9 & 81.3 & 89.7 & 87.6 & 83.0 & 35.1 & 77.7 \\
    \midrule
        \multicolumn{12}{l}{\textbf{External Baselines}} \\
    \midrule
        \,\,OpenAI GPT & 3072 & 100\% & 56.0 & 82.3 & 80.0 & 91.3 & 81.4 & 87.4 & 70.3 & 78.8 & 45.4 & 75.1 \\
        \,\,DistilBERT & 3072 & 100\% & 59.9 & 87.5 & 86.9 & 91.3 & 82.2 & 89.2 & 71.3 & 81.2 & 52.1 & 77.6 \\
        \,\,BERT-base & 3072 & 100\% & 66.4 & 88.9 & 85.8 & 93.5 & 83.4 & 90.5 & 71.2 & 83.0 & 51.3 & 79.6 \\
    \bottomrule 
    
    \end{tabular}
}

\caption{
    The results of various language models on the GLUE-dev test sets.
    $N_{\text{T}}$ denotes the number of neurons available for training, $N_{\text{I}} / N_{\text{T}}$ the proportion of neurons that are used for a single inference.
    ``Avg'' denotes the average score of all the task results to the left of the column.
    \bw{Emphasis} marks the best crammed 1-day UltraFastBERT performance for the given column.
    OpenAI GPT, DistilBERT, and BERT-base refer to models reported in \citet{radford2018improving,sanh2019distilbert,devlin2018bert}.
}
\label{table:downstream_results}
\end{table*}

\section{Model}
\label{section:model}

\subsection{Architecture}
\label{section:architecture}
Our architectural starting point is the crammedBERT architecture \citep{geiping2023cramming}, which we implement to the letter in all but the nature of intermediate layers.
There, the feedforward networks contained in the intermediate layers of the crammedBERT transformer encoder are replaced with fast feedforward networks \cite{belcak2023fast}.

We make the following simplifying changes to the original fast feedforward networks:
\begin{enumerate}
    \item \textit{Remove all differences between leaf and non-leaf nodes.} In particular, we use the same (GeLU) activation function across all nodes, equip all nodes with output weights, and remove all output biases.
    \item \textit{Fix the leaf size to 1.}
    \item \textit{Allow multiple FFF trees in parallel.} We allow for multiple FFF trees to jointly compute the intermediate layer outputs.
    This is achieved by summing the outputs of the individual trees and presenting the sum as the intermediate layer output.
\end{enumerate}
 We denote a model with $K$ trees of depth $D+1$ by appending a suffix to the model name, i.e. UltraFastBERT-$K$x$D$. Note that for consistency with our inference code, we consider a tree with no edges to have depth $0$ -- hence the tree with maximum depth $D$ has depth $D+1$. 
A BERT-base-sized model with the traditional feedforward layer of width 3072 is then just a special case of UltraFastBERT, namely UltraFastBERT-3072x0.

While we share only our fastest model, we train a full range of increasingly deeper and narrower models, starting from UltraFastBERT-3072x0 and proceeding with UltraFastBERT-1536x1, UltraFastBERT-512x2, etc..

\subsection{Training}
\label{section:training}
We follow the final training procedure of crammedBERT \citep{geiping2023cramming}, namely disabling dropout in pretraining and making use of the 1-cycle triangular learning rate schedule.
By default, we train every model for 1 day on a single A6000 GPU, except for the final UltraFastBERT-1x11-long model, which we train 2 times longer using the same regime for slightly better downstream performance.

\subsection{Downstream Performance}
\label{section:performance}

\subsubsection{Setup}
We finetune all UltraFastBERT models for the RTE, MRPC, SST, STS-B, MNLI, QQP, QNLI, and CoLA tasks of the GLUE benchmark \citep{wang2018glue} and report evaluation scores as in \citet{geiping2023cramming} for consistency.
In short, this approach amounts to finetuning for 5 epochs with learning rate $4 \times 10^{-5}$ across all tasks. 

We find that UltraFastBERT models finetuned in this manner for CoLA end up being undertrained if only 5 training epochs are used.
Therefore, we extend the number of CoLA finetuning epochs to 15. This leads to little to no improvement for the baseline crammedBERT models but has a significant impact on the CoLA performance of UltraFastBERTs.

\subsubsection{Results}
The results of our finetuning are listed in \Cref{table:downstream_results}.

We see that UltraFastBERT variants trained for 1 day on a single A6000 GPU all retain at least 96.0\% of the GLUE downstream predictive performance of the original BERT-base model \cite{devlin2018bert}.
We also observe that the performance decreases with the increasing depth of the FFFs.
Note, however, that the majority of the performance decrease due to the increasing depth is caused by only a single task -- CoLA.
This behaviour has previously been observed in the literature and is in line with other work trying to compress BERT behaviour into smaller models \citep{sun2019patient,turc2019well,mukherjee2021xtremedistiltransformers}.
If we disregard CoLA, at least 98.6\% of the predictive performance is preserved by all UltraFastBERT model.

Furthermore, we see that save from CoLA, our best model -- UltraFastBERT-1x11-long -- performs on par with the original BERT-base model while using only 0.3\% of its own neurons, which amounts to a mere 0.4\% of BERT-base neurons. We make the weights of this model public.

\begin{table*}[t!]
\centering
\scalebox{1.07}{
    \begin{tabular}{l|r|rrr|rrr}
    
    \toprule
      \multicolumn{1}{c}{  }
      & \multicolumn{1}{c}{  }
      & \multicolumn{3}{|c}{ CPU Implementation }
      & \multicolumn{3}{|c}{ GPU Implementation } \\
    \midrule

    Model &
    Limit &
    \multirow{1}{*}{Level 1} &
    \multirow{1}{*}{Level 2} &
    \multirow{1}{*}{Level 3} &
    \multirow{1}{*}{Native fused} &
    \multirow{1}{*}{Pytorch BMM} &
    \multirow{1}{*}{Naive CUDA} \\

    \midrule
        BERT-base-4095 & 1.00x & 1.00x & 1.00x & 1.00x  & 1.00x & 1.00x & 1.00x \\
        BERT-base-3072 & 1.33x & 1.55x & 1.74x & 1.39x & 1.33x  & 1.61x &  1.82x    \\
        UltraFastBERT-1x11  & \bw{341.25x}  & \bw{130.7} & \bw{255.1} & -  & -   & \bw{39.45x}& \bw{117.83x}\\        
    \bottomrule 
    
    \end{tabular}
}

\caption{
    The results of the inference acceleration evaluation.
    \bw{Emphasis} highlights the best ``fair comparison'' performance.
}
\label{table:speed_results}
\end{table*}

\begin{algorithm}[h!]
  \KwIn{$B\times H$ input matrix $I$,\hspace{40pt} $(2^D-1)\times H$ weight matrix $W^{\text{in}}$,\,\,\,\,\,\, $(2^D-1)\times H$ weight matrix $W^{\text{out}}$}
  \Intermediate{$B \times D$ logit matrix $L$,\hspace{60pt} $B \times D$ node index matrix $N$}
  \KwOut{$B\times H$ matrix $O$}
  \SetKwFunction{FT}{$\textsc{CMM}$}
  \SetKwProg{Fn}{Function}{:}{}
  \;
  \Fn{\FT{$I, W^{\text{in}}$}}{
    \For{$d \in \left\{1, \dotsc, D-1\right\}$} {
        $L_{\star,d} \gets I\left( W^{\text{in}}_{\left[N_{\star,d-1}\right],\star} \right)^{\text{T}}$\;
        $N_{\star,d} \gets 2N_{\star,d-1} + 1 + \left(L_{\star,d} > 0\right)$\;
    }
    \KwRet{$L, N$}\;
  }
  \;
  \SetKwFunction{FI}{$\textsc{FFF}_I$}
  \SetKwFunction{ACT}{$\textsc{Activation}$}
  \SetKwProg{Fn}{Function}{:}{}
  \Fn{\FI{$I, W^{\text{in}}, W^{\text{out}}$}}{
        $L, N \gets $ \FT{$I,W^{\text{in}}$}\;
        $L \gets \ACT(L)$\;
        \For{$d \in \left\{0, \dotsc, D-1\right\}$} {
            $O_{\star,d} \gets L_{\star,d} \cdot W^{\text{out}}_{N_{\star, d},\star}$\;
        }
        \KwRet{$O$}\;
  }
  \caption{FFF inference forward pass.}
  \label{algorithm:fff}
\end{algorithm}

\section{Inference}
\label{section:inference}
If the purpose of the above part was to report the finding that only very few neurons are needed per inference, it is the goal of this section to adopt the engineering perspective and outline how this can be taken advantage of on the implementation front.

Fast feedforward networks as a part of large language models have a huge acceleration potential.
To indicate the sort of speedup ballpark one could hope for, take GPT-3 \cite{brown2020language}, the first large language model widely lauded for the plausibility of its outputs.
The feedforward networks of each transformer layer of GPT-3 consist of 49152 neurons. If trainable, this network could be replaced with a fast feedforward network of maximum depth 15, which would contain 65536 neurons but use only 16 for inference.
This amounts to about \textbf{0.03\%} of GPT-3's neurons.

At the center of this promise sits the operation of conditional matrix multiplication, with its pseudocode given below, and with our future efforts focused on its efficient implementation.

\subsection{Algorithm}
\label{section:inference_algorithm}
\citet{belcak2023fast} gives recursive pseudocode for FFF inference.
We list the pseudocode for CMM and the consecutive inference for FFFs, with modifications as per \Cref{section:architecture}.
In \Cref{algorithm:fff}, $B$ denotes the batch size, $H$ the layer input width (transformer hidden dimension), $2^D -1$ is the number of neurons, and $M_{\star,k},M_{l,\star}$ denote the $k$-th column and $l$-th row of $M$, respectively. The result of the $>$-comparison in CMM is assumed to be an integer $\in\left\{0,1\right\}$.

\subsection{Compatibility}
\label{section:compatibility}

One may ask whether the conditionality introduced by the use of CMM does not make FFFs incompatible with the processes and hardware already in place for dense matrix multiplication and deep learning more broadly.
In short, the answer is ``No, it does not, save for some increased caching complexity.''

Single-threaded \textit{CPU} DMM as a part of feedforward inference relies on sequential execution of multiplication and accumulation (MAC) instructions.
As such, CPUs, especially edge CPUs, stand to benefit the most easily from the replacement of DMM with CMM as seen in UltraFastBERT, simply because fewer executions of the per-element MAC instructions are needed to compute layer output.
In spite of the apparent use of conditionality, which is commonly associated with branching in CPU code, the ``neural branching'' seen in CMM manifests itself only as an addition of a memory offset to the relevant pointers.
Hence, instruction branch prediction is never engaged to facilitate CMM conditionality.
In order to make full use of weight caching to speed up the access to weights, the CPU might need to be hinted to load only relevant columns of the weight matrix and only one at a time.
Since CMM continues to perform row-column dot products, vector single-instruction-multiple-data (SIMD) parallel processing remains a viable option for speeding up device-specific inference implementations.

The implicitly multi-threaded \textit{GPU} DMM computation makes extensive use of the single-instruction-multiple-threads (SIMT) approach behind modern GPUs by executing the same MAC instructions in each thread, just on different patches of the matrices.
As above, note that this readily carries over to CMM since the conditionality represented by proceeding to different columns of the weight matrices affects only the offset to the memory used, and not which, if, or how many times the MAC instructions are executed.
Nevertheless, efficient DMM implementations distribute the matrix multiplication workload (the pairs of matrix patches to be multiplied) in a manner that maximizes the use of distributed cache so that the accesses to the global device memory, being significantly slower than accessing cache, are limited.
To achieve its full potential with respect to the DMM baseline, any efficient implementation of CMM has to explicitly manage its caching in a way that is optimal for tree traversal, and not patched dense matrix multiplication.
This can be done by always pre-loading the weights of the relevant sub-trees or by using DMM patching strategies but discarding intermediate results from the results of patch margins where not needed.
Either way, it remains to be a challenge to make these optimizations without intimate (and often confidential) knowledge of the implementation's target device.

\subsection{Inference Performance}
\label{section:downstream_performance}

We compare the speed of several available FF/FFF inference implementations.

\paragraph{Implementations.}
For CPU inference, we use the Math Kernel Library available as a part of the Intel oneAPI.
\begin{itemize}
    \item \textbf{Level 1} implementation is the implementation constructed using only BLAS Level 1 routines and BLAS-like Level 1 extensions, namely the vector-vector dot product and scalar-vector product.
    \item \textbf{Level 2} implementation uses batched BLAS Level 2 routines and BLAS-like Level 1 extensions, namely the batched matrix-vector multiplication and batched scalar-vector product.
    \item \textbf{Level 3} implementation uses the (non-batched) BLAS Level 3 matrix-matrix multiplication.
    This is the fastest CPU implementation for FF, but no such implementation can be provided at this time for FFF due to the vector-level sparsity of CMM not being supported by the library.
\end{itemize}

For the GPU implementations, we use either PyTorch kernels or custom CUDA kernels.
\begin{itemize}
    \item \textbf{Native fused} implementation uses the native fused feedforward layer kernel.
    Note that this is the fastest GPU implementation for FF layers but again, no such kernel currently exists for FFFs due to the nature of CMM.
    \item \textbf{BMM} implementation uses the batched matrix multiplication and activation kernels for both FFs and FFFs. In the case of FFFs, we extensively use vector copying at each step of tree descent to simulate conditionality.
    \item \textbf{Naive CUDA} implementation is our custom CUDA kernel code for both FFs and FFFs, performing fused DMM/CMM and activation on the level of vector/matrix elements, executed as a PyTorch extension.
\end{itemize}

\paragraph{Methodology.}
For CPU inference, we perform 250 forward passes per entry on Intel(R) Core(TM) i7-6700HQ CPUs under Intel MKL v2023.2.0, using 64-bit variants of all routines.
We report the mean time taken by single inference, noting that the value of the standard deviation always lay well under 2\% of the mean.
For GPU inference, we perform 1000 forward passes per entry on NVIDIA RTX A6000 GPUs under CUDA v11.7 and PyTorch 2.0.1.
We measure the GPU time and report the mean time taken, with the standard deviation again well under 2\% of the mean in all cases.
We take batch size $B=128 \times 128$ (equivalent to the BERT pretraining context token batch size) and hidden dimension $H=768$.

\paragraph{Results.}
\Cref{table:speed_results} lists the performance comparison of feedforward and fast feedforward layers as they appear in BERT-base and UltraFastBERT-1x11.
Each column of the table lists the relative inference FFF-over-FF implementation speedups \textit{when using the same linear-algebraic routine primitives}.

The two entries missing \Cref{table:speed_results} are for the currently unavailable BLAS Level 3 and Native fused implementations of FFFs.

\paragraph{Further comparisons.}
All of the speedups reported in \Cref{table:speed_results} give ``fair comparisons'', meaning that in each case, both the FF and FFF implementation used exactly the same primitive linear-algebraic operations.
One may also be interested in knowing how the best implementations of FFF currently fair against the best implementations of FF, even though the ones for FF use primitives unavailable for FFF.
On CPU, the Level 1 and Level 2 implementations of FFF perform inference \textbf{48x and 78x} faster than the fastest (Level 3) implementation of FF, respectively.
On GPU, the PyTorch BMM implementation of FFF delivers a \textbf{3.15x} speedup over the fastest (Native fused) implementation of FF.

\subsection{Future outlook}
\label{section:future_outlook}

The broad strokes for starting efficient implementation of FFF inference have already been painted as a part of the PyTorch library.
Hybrid vector-level sparse tensors, if fully supported for singular and batched matrix multiplication, would suffice to implement CMM and FFF inference as in \Cref{algorithm:fff}.

A further native implementation of CMM as a part of device-specific Intel MKL/NVIDIA cuBLAS code would stand a real chance of fully delivering on the promise of 341-fold speedup.

\section{Conclusion}
\label{section:conclusion}

We present UltraFastBERT, a modified version of the (crammed)BERT architecture that uses fast feedforward instead of feedforward networks in its intermediate layers.
UltraFastBERT serves as proof that large language models only really need to engage an exponential fraction of their parameters to perform individual inferences.
UltraFastBERT-1x11, our deepest model with the highest promise of acceleration, uses only 0.3\% of its neurons during inference and already achieves a 78x CPU speedup over the inference time of the corresponding feedforward layer.
With a theoretical speedup promise of 341x at the scale of BERT-base models, we hope that our work will inspire an effort to implement primitives for conditional neural execution as a part of device programming interfaces.  

\bibliography{paper}

\begin{thebibliography}{11}
\providecommand{\natexlab}[1]{#1}
\providecommand{\url}[1]{\texttt{#1}}
\expandafter\ifx\csname urlstyle\endcsname\relax
  \providecommand{\doi}[1]{doi: #1}\else
  \providecommand{\doi}{doi: \begingroup \urlstyle{rm}\Url}\fi

\bibitem[Anil et~al.(2023)Anil, Dai, Firat, Johnson, Lepikhin, Passos, Shakeri, Taropa, Bailey, Chen, et~al.]{anil2023palm}
Anil, R., Dai, A.~M., Firat, O., Johnson, M., Lepikhin, D., Passos, A., Shakeri, S., Taropa, E., Bailey, P., Chen, Z., et~al.
\newblock Palm 2 technical report.
\newblock \emph{arXiv preprint arXiv:2305.10403}, 2023.

\bibitem[Belcak \& Wattenhofer(2023)Belcak and Wattenhofer]{belcak2023fast}
Belcak, P. and Wattenhofer, R.
\newblock Fast feedforward networks.
\newblock \emph{arXiv preprint arXiv:2308.14711}, 2023.

\bibitem[Brown et~al.(2020)Brown, Mann, Ryder, Subbiah, Kaplan, Dhariwal, Neelakantan, Shyam, Sastry, Askell, et~al.]{brown2020language}
Brown, T., Mann, B., Ryder, N., Subbiah, M., Kaplan, J.~D., Dhariwal, P., Neelakantan, A., Shyam, P., Sastry, G., Askell, A., et~al.
\newblock Language models are few-shot learners.
\newblock \emph{Advances in neural information processing systems}, 33:\penalty0 1877--1901, 2020.

\bibitem[Devlin et~al.(2018)Devlin, Chang, Lee, and Toutanova]{devlin2018bert}
Devlin, J., Chang, M.-W., Lee, K., and Toutanova, K.
\newblock Bert: Pre-training of deep bidirectional transformers for language understanding.
\newblock \emph{arXiv preprint arXiv:1810.04805}, 2018.

\bibitem[Geiping \& Goldstein(2023)Geiping and Goldstein]{geiping2023cramming}
Geiping, J. and Goldstein, T.
\newblock Cramming: Training a language model on a single gpu in one day.
\newblock In \emph{International Conference on Machine Learning}, pp.\  11117--11143. PMLR, 2023.

\bibitem[Mukherjee et~al.(2021)Mukherjee, Awadallah, and Gao]{mukherjee2021xtremedistiltransformers}
Mukherjee, S., Awadallah, A.~H., and Gao, J.
\newblock Xtremedistiltransformers: Task transfer for task-agnostic distillation.
\newblock \emph{arXiv preprint arXiv:2106.04563}, 2021.

\bibitem[Radford et~al.(2018)Radford, Narasimhan, Salimans, Sutskever, et~al.]{radford2018improving}
Radford, A., Narasimhan, K., Salimans, T., Sutskever, I., et~al.
\newblock Improving language understanding by generative pre-training.
\newblock 2018.

\bibitem[Sanh et~al.(2019)Sanh, Debut, Chaumond, and Wolf]{sanh2019distilbert}
Sanh, V., Debut, L., Chaumond, J., and Wolf, T.
\newblock Distilbert, a distilled version of bert: smaller, faster, cheaper and lighter.
\newblock \emph{arXiv preprint arXiv:1910.01108}, 2019.

\bibitem[Sun et~al.(2019)Sun, Cheng, Gan, and Liu]{sun2019patient}
Sun, S., Cheng, Y., Gan, Z., and Liu, J.
\newblock Patient knowledge distillation for bert model compression.
\newblock \emph{arXiv preprint arXiv:1908.09355}, 2019.

\bibitem[Turc et~al.(2019)Turc, Chang, Lee, and Toutanova]{turc2019well}
Turc, I., Chang, M.-W., Lee, K., and Toutanova, K.
\newblock Well-read students learn better: On the importance of pre-training compact models.
\newblock \emph{arXiv preprint arXiv:1908.08962}, 2019.

\bibitem[Wang et~al.(2018)Wang, Singh, Michael, Hill, Levy, and Bowman]{wang2018glue}
Wang, A., Singh, A., Michael, J., Hill, F., Levy, O., and Bowman, S.~R.
\newblock Glue: A multi-task benchmark and analysis platform for natural language understanding.
\newblock \emph{arXiv preprint arXiv:1804.07461}, 2018.

\end{thebibliography}
\bibliographystyle{icml2023}



\end{document}